\documentclass{article}

\usepackage[dblblindworkshop, final]{neurips_2026}
\workshoptitle{1st NeurIPS Workshop on Agentic Web: Decentralized, Continually-Adapting Agent Ecosystems}

\usepackage[utf8]{inputenc}
\usepackage[T1]{fontenc}
\usepackage{hyperref}
\usepackage{url,color}
\usepackage{booktabs}
\usepackage{amsfonts}
\usepackage{amsmath}
\usepackage{nicefrac}
\usepackage{microtype}
\usepackage{xcolor}
\usepackage{graphicx}

\title{Hardening Agentic Web Populations Through Multi-Criterion Competitive Co-Evolution: Lessons from Wargame Strategy Optimization}

\author{Shashank Raj \\
Department of Computer Science \\
The University of Alabama \\
Tuscaloosa, AL 35487, USA \\
\texttt{sraj2@crimson.ua.edu}
\And
Kalyanmoy Deb \\
Department of Computer Science \\
The University of Alabama \\
Tuscaloosa, AL 35487, USA \\
\texttt{kdeb@ua.edu}
}

\begin{document}
\maketitle

\begin{abstract}
Agents on the open web will face adversaries that keep adapting, so a defense that is only tested once will eventually be found and broken. We propose treating jailbreak attacks and safety guardrails as two populations that evolve against each other, rather than as a fixed attacker probing a fixed defense. This idea comes directly from wargame strategy optimization, where two-population competitive co-evolution with alternating evolution cycles and cross-play evaluation has already been shown to produce strategies that are robust to a moving opponent, and where a companion line of work has shown that the same machinery can measure how much competitive pressure a population must absorb. However, real-world applications can never rely on a single criterion as a key performance indicator, rather must use multiple conflicting criteria, such as, cost and efficiency. We describe how this mechanism transfers to language model safety, argue why the resulting multi-criteria, competitive and coevolutionary arms race is a genuinely networked, population-level problem rather than a two-agent toy setting, and lay out the open problems that stand between the wargame version of this idea and an agentic web version of it.
\end{abstract}

\section{Introduction}
\label{sec:intro}

Most evaluations of language model safety still follow a single-shot pattern: a fixed set of jailbreak prompts is run against a fixed model, once. This misses the central fact about the agentic web: attackers and defenders are not static. An attacker who fails against today's guardrail will keep adapting until a new prompt works, and a defender who only ever sees yesterday's attacks will be blind to tomorrow's. Static red-teaming benchmarks cannot capture this back-and-forth, because they freeze one side of the interaction.

\begin{figure}[hbt]
  \centering
  \includegraphics[width=\linewidth]{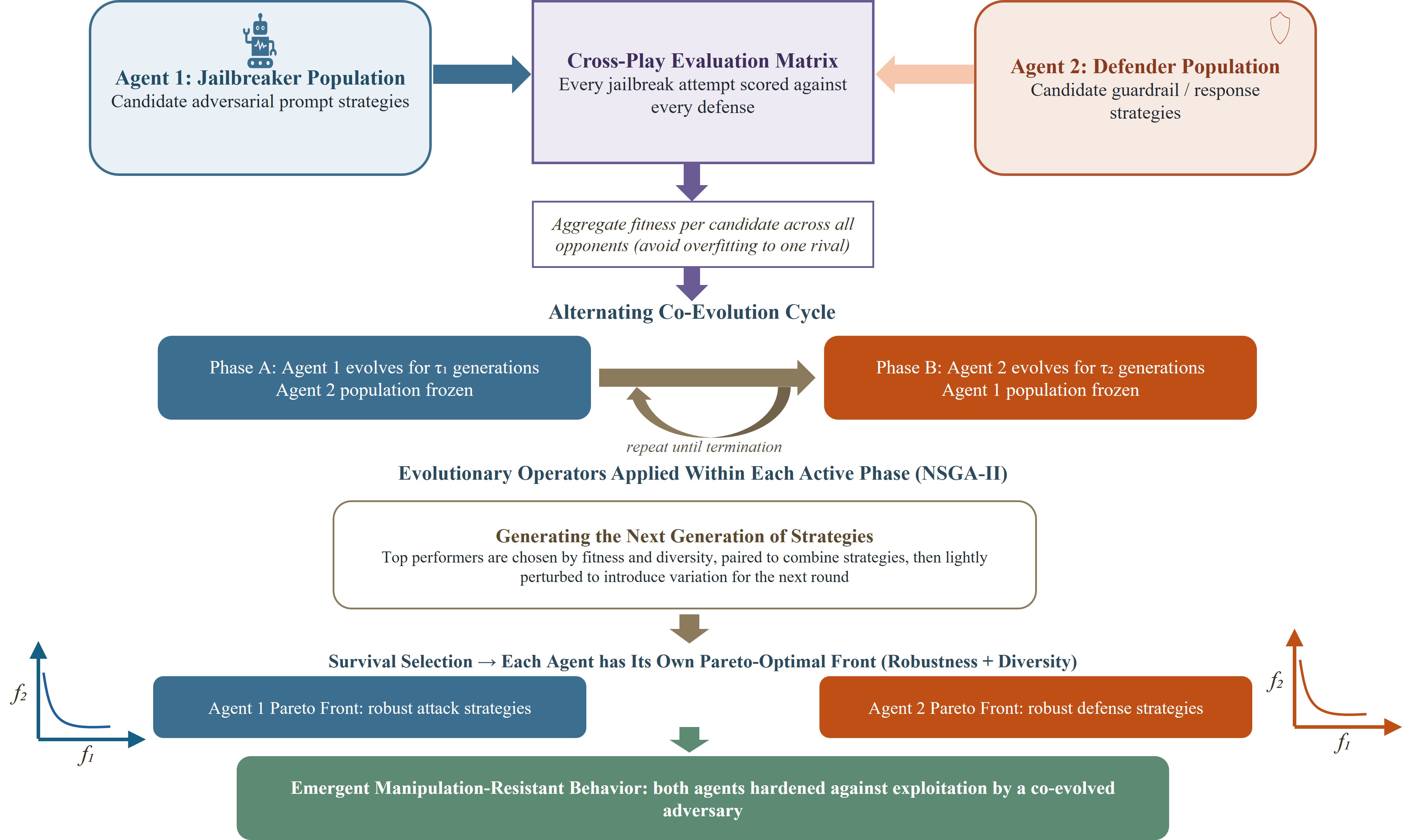}
  \caption{Our proposed mechanism, adapted from multi-criterion competitive co-evolution in wargame strategy optimization. A jailbreaker population and a defender population are scored against each other in a cross-play matrix, fitness vectors are aggregated across all opponents to avoid overfitting to a single rival, and the two populations take turns evolving while the other is held fixed. Each side keeps its own Pareto-optimal set of solutions resulting from multiple conflicting criteria, and the intended outcome is that both sides become resistant to manipulation by a co-evolved adversary rather than by a fixed one.}
  \label{fig:sketch}
\end{figure}

We propose a different setup, illustrated in Figure~\ref{fig:sketch}: run the attacker and the defender as two co-evolving populations, and let them adapt to each other over many rounds instead of one. This is not a new idea in coevolutionary computation. It is the same mechanism used to optimize offense and defense strategies in wargame simulations \citep{guha2024attacker,guha2025progressive,guha2025regularity}, and it is the same mechanism a companion line of work has used to measure competitive balance directly on a simulated game \citep{raj2026balance,raj2026coin}. Our contribution here is a position, not a new algorithm: we argue that this two-population, alternating, cross-play mechanism is a natural fit for hardening agentic web systems against external manipulation, and we describe what has to change to move it from a wargame simulator to a population of language model agents interacting at web scale.

\section{Background: Competitive co-evolution in wargame strategy optimization}
\label{sec:background}

The mechanism in Figure~\ref{fig:sketch} was developed for a different domain: optimizing offense and defense strategies in a military simulation. \citet{guha2024attacker} formulated this as a two-population, multi-criterion problem. An offense population and a defense population are each made of candidate strategies, and instead of evaluating a strategy against a single opponent, every strategy is evaluated against every member of the opposing population. This is the cross-play evaluation matrix in Figure~\ref{fig:sketch}. Formally, if $x$ is an offense strategy and $y$ is a defense strategy, a single match produces an objective vector
\begin{equation}
\mathbf{f}(x, y) = [f_1^{\text{off}}(x, y), f_2^{\text{off}}(x, y), f_1^{\text{def}}(x, y), f_2^{\text{def}}(x, y)]^T,
\end{equation}
and the full set of matches between an offense population of size $N_{\text{off}}$ and a defense population of size $N_{\text{def}}$ produces an evaluation tensor $\mathbf{F}^t \in \mathbb{R}^{N_{\text{off}} \times N_{\text{def}} \times 4}$ at generation $t$. Each strategy's fitness is obtained by aggregating its row or column of this tensor across all opponents it faced, rather than by its result against any one opponent. This aggregation step is what prevents a strategy from winning simply by overfitting to a single rival, which is a well known failure mode of competitive co-evolution \citep{antonio2017survey}.

The two populations do not evolve at the same time. Instead, following an alternating cycle $[\tau_{\text{off}}, \tau_{\text{def}}]$, one population evolves for $\tau_{\text{off}}$ generations using NSGA-II \citep{deb2002nsga2} while the other is held fixed, then the roles switch. This alternation, together with the cross-play aggregation, is what produces populations whose Pareto-optimal fronts represent strategies robust to an adapting opponent rather than to a single frozen one. Follow-up work extended this framework with progressive shrinking of the decision space for more realistic sequential decision-making \citep{guha2025progressive}, and with a regularity-based procedure for selecting one actionable strategy off the Pareto front \citep{guha2025regularity}.

A companion line of work, \citet{raj2026balance} and \citet{raj2026coin}, applies the same two-population, alternating, cross-play mechanism to a different question: not just optimizing strategy, but measuring balance. That work introduces a movement metric that compares a hypervolume \cite{lyndon,bader-deb-ecki} achieved under a cooperative upper bound (a strategy that assumes a fully cooperative opponent) against the hypervolume actually achieved at co-evolutionary equilibrium. The gap between these two,
\begin{equation}
\text{Movement} = \text{HV}_{\text{coop}} - \text{HV}_{\text{coev}},
\end{equation}
quantifies how much a side is forced to sacrifice by competitive pressure. When both sides have approximately equal movement, the interaction is judged balanced; when one side's movement is much larger than the other's, that side is systematically disadvantaged. This matters for our proposal because it shows that the same co-evolutionary machinery used to harden strategies can also be turned into a diagnostic: it does not just produce robust attackers and defenders, it can tell us how much pressure each side is actually under.

\section{Mapping the wargame mechanism to jailbreak and defense coevolutionary framework}
\label{sec:mapping}

The translation from wargame agents to language model agents is direct. The offense population becomes a population of jailbreaker strategies: candidate prompts, prompt templates, or prompting policies whose objective is to elicit a disallowed response while remaining hard to detect. The defense population becomes a population of guardrail strategies: system prompts, classifiers, or refusal policies whose objective is to block disallowed responses while keeping the false refusal rate low. Both sides face a genuine multi-criterion trade-off, which is why the multi-criterion machinery in Section~\ref{sec:background} carries over rather than a simple win or lose signal: an attacker that always succeeds but is trivially detectable is not actually strong, and a defender that never fails but refuses half of all benign requests is not actually useful. The existing studies either do not consider multiple objectives for each agent in coevolutionary studies or combine multiple criteria into a single one by a scalarization approach which are always subjective and non-generic. 

The part of the mechanism we think matters most is the combination of cross-play aggregation and alternation. Existing adversarial robustness work for language models typically evaluates a fixed set of jailbreak techniques against a fixed model \citep{perez2022redteam,zou2023universal,chao2023pair}, which finds real weaknesses but only tells us about that particular snapshot of attacker and defender. If a defender is instead only trained against the current best-known attack, it can overfit to that attack and remain fragile against a different one, and the reverse is true for an attacker trained against a single defense. Cross-play evaluation directly addresses this: a candidate jailbreaker is scored against the entire defender population, not the single best defender, and its fitness is an {\em aggregate\/} measure of all of them. Alternation between the two phases then makes sure that when the defender population is updated, the jailbreaker population has to keep up in the next phase rather than the two populations drifting independently. Importantly, the multiple Pareto strategies provide alternative options to each agent to analyze and choose a single preferred one. With a single criterion based coevolutionary approach will always produce a single strategy for each agent at each step, without the flexibility of choosing among a few by the human or machine-based agents, forcing agents to "take" it or "leave" it.

We can also carry over the movement metric idea in Section~\ref{sec:background} as a diagnostic for safety, not just as a robustness training procedure. A cooperative upper bound for the jailbreaker side would be its attack success rate against an undefended model; a cooperative upper bound for the defender side would be its refusal accuracy against a fully compliant, non-adversarial user. The gap between each of these upper bounds and what each side actually achieves at co-evolutionary equilibrium is a direct, measurable answer to a question that is currently answered mostly by intuition: how much is a given guardrail actually costing an attacker, and how much is a given attacker actually costing a defender's usability, providing a transparent trade-off information at every step to both agents.

\section{Why the proposed mapping is an agentic web problem?}
\label{sec:agenticweb}

The setup above is still a two-population picture, which on its own looks like an adversarial training method rather than something specific to the agentic web. We think the extension to that setting is what makes this workshop the right venue for this study, for three reasons.

First, at internet scale the attacker and defender sides are not each a single strategy space, they are themselves heterogeneous populations of independently operated agents that discover each other rather than being paired by a benchmark designer. A discovery and pairing mechanism, not just an evaluation loop, becomes part of the problem: which jailbreaker agents get matched against which defender agents, and how that matching itself evolves, is closely related to the capability discovery question this workshop already identifies as open.

Second, credit assignment becomes harder once there are many agents on each side rather than two populations. In the wargame version, aggregating a strategy's fitness across every opponent it faced is tractable because both populations are of a fixed, moderate size and every match is a cheap simulation. At web scale, a defender agent may be probed by thousands of independently evolving attacker lineages, and a single simulation is not a cheap local computation but an actual language model call. Whether alternating, cross-play co-evolution still converges rather than destabilizes when the population sizes and the number of simultaneous lineages grow is an open question, and it is exactly the kind of learning-dynamics question we think this workshop is targeting to answer.

Third, the outcome we want is not just two robust populations in isolation, it is a form of trust that can propagate across a network: an agent that has proven resistant to a wide, co-evolved range of manipulation should be able to carry that reputation into interactions with agents it has never met. This connects the mechanism directly to the manipulation-resistant reputation problem this workshop names as a core open direction, rather than leaving it as a self-contained adversarial training result.

\section{Open problems and a note on dual use}
\label{sec:openproblems}

Moving this mechanism from a wargame simulator to language model agents raises problems that did not exist in the original setting. The cross-play evaluation matrix requires $N_{\text{off}} \times N_{\text{def}}$ evaluations per generation; when each evaluation is a language model call rather than a fast simulator step, this cost grows quickly and some form of surrogate modeling or opponent sub-sampling, similar in spirit to the surrogate models used in the earliest wargame version of this framework \citep{guha2024attacker}, will likely be necessary. A more recent version of this framework removes surrogates entirely and operates directly on the game dynamics to avoid surrogate inaccuracy \citep{raj2026coin}, which is only possible because a wargame simulator step is cheap; a language model call is not, so this trade-off between surrogate accuracy and evaluation cost will need to be revisited for agentic web settings. 
The two-population framework also makes a  simplification: a more realistic agentic web setting has many attacker and defender lineages rather than just two agents, and it is not yet known whether the alternating cycle that keeps two populations from drifting apart still works once there are many populations updating on different schedules. In the presence of multiple agents, how would the sequence of agents affect the final optimal strategies must be researched. Moreover, the effect of cooperation and/or conflict among multiple jailbreaker agents on defender's optimal strategy would be another interesting direction to pursue. Our multi-criterion coevolutionary strategy is generic enough to study all these possibilities, a gamut of which we plan to embed in an agentic web application. 

Finally, we want to be direct about their dual use. A method for co-evolving strong jailbreak strategies is, by construction, also a method for producing strong jailbreak strategies, and any implementation of this idea should keep the co-evolved attacker population access-controlled and should report defender robustness rather than attacker prompts.

\section{Conclusion}
\label{sec:conclusion}

Competitive co-evolution with alternating cycles and cross-play evaluation in the original study was built  to optimize and to measure balance between two adapting sides in a wargame setting %\citep{guha2024attacker,guha2025progressive,guha2025regularity,raj2026balance,raj2026coin}. 
We have argued here that the same mechanism is a natural way to harden jailbreaker and defender agents against each other. The ability to consider multiple conflicting decision criteria for each agent is an unique hallmark feature of our approach. The multi-criterion coevolutionary approach allows each agent to analyze multiple trade-off Pareto strategies before choosing the most preferred and appropriate strategy in every move, thereby providing the confidence and comprehensiveness of chosen decisions. Extending it to many heterogeneous, discoverable agents at web scale, rather than two fixed populations, is what turns it from an adversarial training trick into a genuine agentic web research problem.

\bibliographystyle{plainnat}
\bibliography{references}

@inproceedings{bader-deb-ecki,
author={J. Bader and K. Deb and E. Zitzler},
title={Faster Hypervolume-based Search using {M}onte {C}arlo Sampling},
booktitle={Proceedings of Multiple Criteria Decision Making (MCDM 2008)}, 
note={LNEMS-634}, pages={313--326}, publisher={Heidelberg: Springer}, year={2010}}

@article{lyndon,
journal={IEEE Transactions on Evolutionary Computation}, volume={10}, number={1}, year={2006}, pages={29--38},
title={A Faster Algorithm for Calculating Hypervolume},
author={L. While and P. Hingston and L. Barone and
S. Huband}}

@inproceedings{guha2024attacker,
  title={Attacker-defender strategy optimization using multi-objective competitive co-evolution},
  author={Guha, Ritam and Mckendrick, Ryan and Feest, Bradley and Deb, Kalyanmoy},
  booktitle={International Conference on Parallel Problem Solving from Nature},
  pages={351--366},
  year={2024},
  organization={Springer}
}

@inproceedings{guha2025progressive,
  title={A Multi-objective Competitive Co-evolutionary Framework with Progressive Shrinking for Wargame Scenarios},
  author={Guha, Ritam and McKendrick, Ryan and Feest, Bradley and Deb, Kalyanmoy},
  booktitle={International Conference on Evolutionary Multi-Criterion Optimization},
  pages={288--302},
  year={2025},
  organization={Springer}
}

@article{guha2025regularity,
  title={Multiobjective competitive co-evolutionary optimization and regularity-based decision-making for two-agent wargame strategy optimization},
  author={Guha, Ritam and Mckendrick, Ryan and Feest, Bradley and Deb, Kalyanmoy},
  journal={IEEE Transactions on Evolutionary Computation},
  volume={30},
  number={2},
  pages={652--666},
  year={2025},
  publisher={IEEE}
}

@inproceedings{raj2026balance,
  title={To Balance Competitive Games Using a Multi-objective Coevolutionary Approach},
  author={Raj, Shashank and Garrard, Auden and McKendrick, Ryan and Feest, Bradley and Deb, Kalyanmoy},
  booktitle={Proceedings of the Genetic and Evolutionary Computation Conference},
  pages={1191--1199},
  year={2026}
}

@techreport{raj2026coin,
  title={Towards Balanced Wargames: A Multi-Objective Co-Evolutionary Adversarial Approach for Full-Game and Meta-Game Balance},
  author={Raj, Shashank and McKendrick, Ryan and Feest, Bradley and Deb, Kalyanmoy},
  institution={COIN Report},
  number={2026008},
  year={2026},
  url={https://kdeb.us/papers/c2026008.pdf},
  note={Accessed August 23, 2026}
}

@article{deb2002nsga2,
  title={A fast and elitist multiobjective genetic algorithm: {NSGA-II}},
  author={Deb, Kalyanmoy and Pratap, Amrit and Agarwal, Sameer and Meyarivan, T.},
  journal={IEEE transactions on evolutionary computation},
  volume={6},
  number={2},
  pages={182--197},
  year={2002},
  publisher={Ieee}
}

@article{antonio2017survey,
  title={Coevolutionary multiobjective evolutionary algorithms: Survey of the state-of-the-art},
  author={Antonio, Luis Miguel and Coello, Carlos A Coello},
  journal={IEEE Transactions on Evolutionary Computation},
  volume={22},
  number={6},
  pages={851--865},
  year={2017},
  publisher={IEEE}
}

@article{perez2022redteam,
  title={Red teaming language models with language models},
  author={Perez, Ethan and Huang, Saffron and Song, Francis and Cai, Trevor and Ring, Roman and Aslanides, John and Glaese, Amelia and McAleese, Nat and Irving, Geoffrey},
  booktitle={Proceedings of the 2022 conference on empirical methods in natural language processing},
  pages={3419--3448},
  year={2022}
}

@article{zou2023universal,
  title={Universal and transferable adversarial attacks on aligned language models},
  author={Zou, Andy and Wang, Zifan and Carlini, Nicholas and Nasr, Milad and Kolter, J Zico and Fredrikson, Matt},
  journal={arXiv preprint arXiv:2307.15043},
  year={2023}
}

@inproceedings{chao2023pair,
  title={Jailbreaking black box large language models in twenty queries},
  author={Chao, Patrick and Robey, Alexander and Dobriban, Edgar and Hassani, Hamed and Pappas, George J and Wong, Eric},
  booktitle={2025 IEEE Conference on Secure and Trustworthy Machine Learning (SaTML)},
  pages={23--42},
  year={2025},
  organization={IEEE}
}

\end{document}